# Bisimulations for Fuzzy Transition Systems

Yongzhi Cao, Guoqing Chen, and Etienne Kerre

*Abstract*—There has been a long history of using fuzzy language equivalence to compare the behavior of fuzzy systems, but the comparison at this level is too coarse. Recently, a finer behavioral measure, bisimulation, has been introduced to fuzzy finite automata. However, the results obtained are applicable only to finite-state systems. In this paper, we consider bisimulation for general fuzzy systems which may be infinite-state or infinite-event, by modeling them as fuzzy transition systems. To help understand and check bisimulation, we characterize it in three ways by enumerating whole transitions, comparing individual transitions, and using a monotonic function. In addition, we address composition operations, subsystems, quotients, and homomorphisms of fuzzy transition systems and discuss their properties connected with bisimulation. The results presented here are useful for comparing the behavior of general fuzzy systems. In particular, this makes it possible to relate an infinite fuzzy system to a finite one, which is easier to analyze, with the same behavior.

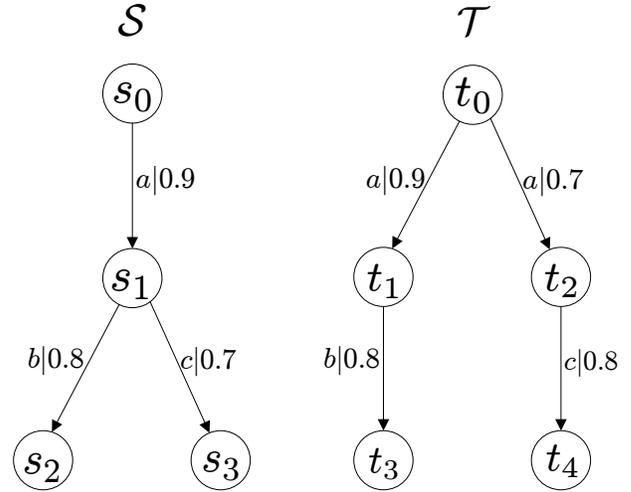

Fig. 1. Fuzzy language equivalent systems may not be bisimilar.

## I. Introduction

THE idea of fuzzy systems was originated by Zadeh in 1965 [42]. One of the main research directions on fuzzy systems is to consider fuzzy systems as a generalization of nondeterministic automata or Petri nets and investigate them within the same conceptual framework as classical systems. As a natural generalization of nondeterministic automata, the mathematical formulation of fuzzy automata was first proposed by Wee in 1967 [38]. The basic idea in the formulation is that, unlike the classical case, a fuzzy automaton can switch from one state to another one to a certain (truth) degree, and thus it is capable of capturing the uncertainty appearing in states or state transitions of a system. In the literature up to now (see, for example, [1], [8], [12], [14], [19], [23], [28], [34], [39], [41]), a great variety of types of fuzzy automata has been proposed in different modeling situations and the notion of fuzzy automata has proved useful in many areas such as learning control and pattern recognition. In parallel, various fuzzy Petri nets have been formulated and extensively investigated (see [4]–[6], [9], [18], [31] and the references therein).

This work was supported by the National Natural Science Foundation of China under Grants 70890080, 60873061, and 60973004 and by the National Basic Research Program of China (973 Program) under Grants 2007CB311003, 2009CB320701, and 2010CB328103.

Y. Cao is with the Institute of Software, School of Electronics Engineering and Computer Science, and the Key Laboratory of High Confidence Software Technologies, Peking University, Beijing 100871, China (e-mail: caoyz@pku.edu.cn).

G. Chen is with the School of Economics and Management, Tsinghua University, Beijing 100084, China (e-mail: chengq@sem.tsinghua.edu.cn).

E. Kerre is with the Department of Applied Mathematics and Computer Science, Ghent University, Krijgslaan 291 (S9), B-9000 Gent, Belgium (e-mail: etienne.kerre@ugent.be).

When modeling a system by fuzzy automata (or fuzzy Petri nets), it is often possible to define multiple models of the same system. Given different fuzzy automaton models of a system, there is a need for some formal techniques that can be used to compare these models. A straightforward idea is to employ the concept of fuzzy language equivalence that stipulates that two fuzzy automata are equivalent if they accept the same strings of input symbols with the identical membership grade [14], [19], [24], [28]. Although this equivalence has been extensively used in both theory and application, it is sometimes considered to be too coarse. For example, the systems $\mathcal{S}$ and $\mathcal{T}$ in Fig. 1 are fuzzy language equivalent if we simply identify every state as being accepting, but their behavior is different: $\mathcal{S}$ can always choose between $b$ and $c$ after performing $a$, while $\mathcal{T}$ can only execute either $b$ or $c$ (but not both) after $a$. More seriously, fuzzy language equivalent systems can have different deadlocking behavior (inability to proceed).

In fact, the above deficiency of fuzzy language equivalence is not due to fuzziness, but to nondeterminism. For classical nondeterministic systems, this matter has been satisfactorily resolved by introducing the important notion of bisimulation [27], [29]. A bisimulation is a binary relation between discrete event systems like process algebras, Petri nets or automata models, associating systems which behave in the same way in the sense that one system simulates the other and vice-versa. Intuitively, two systems are bisimilar if they match each other's moves. Bisimulation equivalence allows one to relate bisimilar systems and to reduce the state space of a system by combining bisimilar states to generate a quotient system with an equivalent behavior but with fewer states. In the past two decades, bisimulation has been considerably extended to probabilistic and stochastic systems (see, for example, [2], [3], [16], [22], [26] and references therein).

However, to our knowledge, few efforts, except the work [3], [32], [36], have been made to consider the bisimulation for fuzzy automata, or more generally, fuzzy systems. Following the algebraic theory of classical automata, Petković introduced the concept of congruence for fuzzy automata in [32]. It turns out that such a congruence is exactly a bisimulation. Based on this concept, an improved minimization algorithm for fuzzy automata has been developed there. Recently, Buchholz has put forward a general definition of bisimulation for weighted automata over a generic semiring which includes well-known automata models as specific cases [3]. A partition refinement algorithm using a matrix notation is provided to compute the largest bisimulation for a given weighted automaton and a definition of "aggregated" automaton, corresponding to a quotient, is presented as well. By instantiating the semiring in the framework of [3] to the closed unit interval $[0,1]$ with binary operations $\max$ and $\min$, Sun *et al.* investigated the forward and backward bisimulations for fuzzy automata [36].

It is worth noting that the methods in [3], [32], [36] are applicable only to finite (finite-state and finite-event) fuzzy systems. It is clearly a limitation, because many fuzzy systems are (or should be seen as) infinite-state or infinite-event. For example, the dynamic fuzzy systems studied in [21], fuzzy discrete event systems modeled by max-product automata [25], and fuzzy stochastic automata [34] involve an infinite number of states and fuzzy automata for computing with all words in [8] require an infinite number of events. This observation motivates us to introduce and explore the concept of bisimulation for general fuzzy systems which may be infinite-state or infinite-event. To this end, we model fuzzy systems as fuzzy transition systems (FTSs) and define bisimulation between them in this paper. An FTS is characterized by a (possibly infinite) set of states, including the initial state, a (possibly infinite) set of labels, and a set of fuzzy transitions. Although many formal description tools for fuzzy systems such as fuzzy Petri nets are not FTSs, it is possible to translate a system's description in one of these formalisms into the FTS representing its behavior.

FTSs are a natural generalization of the widely used formal models—labeled transition systems—in computer science [20]. On the other hand, they are an extension of fuzzy automata by allowing state set and label set to be infinite. FTSs and some probabilistic transition systems [11], [16], [26] seem to be similar, but they have two major differences: One is that when a probabilistic transition system is put into operation, there is no vagueness in the current state, next state, and the extent to which they will be activated. At any time (upon each label), one and exactly one state will be activated with an implied membership value of 1. The other difference is that in a probabilistic transition system, the sum of the weights for some transitions (from the current state) should be one, whilst there is no such requirement for an FTS.

We define the notion of bisimulation between FTSs based on a general binary relation, which is not necessarily an equivalence relation as required in [3], [32], [36] and makes it more convenient for bisimulation checking. Due to the relaxation of relation, we have to consider the so-called correlational pairs in place of equivalence classes in [3], [32], [36]. As a result, the bisimulation equates two FTSs whenever they perform the same action with the same maximum possibility in each correlational pair. We show that the union of all bisimulations between two TFSs gives rise to the largest bisimulation, called bisimilarity. Bisimulation and bisimilarity are also characterized in two other ways: One is to compare individual transitions which does not involve correlational pairs; the other is based upon a suitable monotonic function which has the bisimilarity as the largest fixed point. These are just different formal presentations of the same thing, but they may help in understanding and checking bisimulation and bisimilarity. Further, we address composition operations, subsystems, quotients, and homomorphisms of FTSs and discuss their properties connected with bisimulation. The results presented in the paper are helpful in comparing the behavior of FTSs. In particular, this makes it possible to relate an infinite fuzzy system to a finite one, which is easier to analyze, with the same behavior.

The remainder of this paper is structured as follows. We briefly review some basics of fuzzy sets and introduce the concept of FTSs in Section II. Section III is devoted to the definition of bisimulation using correlational pairs and the existence of bisimilarity. We specialize the notion of bisimulation for equivalence relations and for fuzzy finite automata in this section. In Section IV, we characterize bisimulation and bisimilarity by comparing individual transitions and explore more properties of bisimulation. In Section V, by defining a monotonic function we give a necessary and sufficient condition for a relation to be a bisimulation and show that bisimilarity is exactly the largest fixed point of the function. The notions of subsystems, quotients, and homomorphisms of FTSs and their properties connected with bisimulation are addressed in Section VI. The paper is concluded in Section VII with a brief discussion on the future research.

## II. Fuzzy Transition Systems

In this section, after briefly recalling a few basic facts on fuzzy set theory, we present fuzzy transition systems (FTSs) as a basic model for some fuzzy systems.

Let $X$ be a universal set. A *fuzzy set* $A$, or rather a *fuzzy subset* $A$ of $X$, is defined by a function assigning to each element $x$ of $X$ a value $A(x)$ in $[0,1]$. Such a function is called a *membership function*, which is a generalization of the characteristic function associated to a crisp subset of $X$; the value $A(x)$ characterizes the degree of membership of $x$ in $A$. The *support* of a fuzzy set $A$ is a crisp set defined as $\mathrm{supp}(A) = \{x \in X : A(x) > 0\}$. Whenever $\mathrm{supp}(A)$ is a finite set, say $\mathrm{supp}(A) = \{x_1, x_2, \ldots, x_n\}$, we may write $A$ in Zadeh's notation as

$$A = \frac{A(x_1)}{x_1} + \frac{A(x_2)}{x_2} + \cdots + \frac{A(x_n)}{x_n}.$$

A fuzzy subset of $X$ can be used to formally represent a possibility distribution on $X$. We denote by $\mathcal{F}(X)$ the set of all fuzzy subsets of $X$.

For any $A, B \in \mathcal{F}(X)$, we say that $A$ is contained in $B$ (or $B$ contains $A$), denoted by $A \subseteq B$, if $A(x) \leq B(x)$ for all $x \in X$. We say that $A = B$ if and only if $A \subseteq B$ and $B \subseteq A$.

For any family $\lambda_i$, $i \in I$, of elements of $[0,1]$, we write $\vee_{i \in I} \lambda_i$ or $\vee \{\lambda_i : i \in I\}$ for the supremum of $\{\lambda_i : i \in I\}$, and $\wedge_{i \in I} \lambda_i$ or $\wedge \{\lambda_i : i \in I\}$ for the infimum. In particular, if $I$ is finite, then $\vee_{i \in I} \lambda_i$ and $\wedge_{i \in I} \lambda_i$ are the greatest element and the least element of $\{\lambda_i : i \in I\}$, respectively. For any $\mu \in \mathcal{F}(X)$ and $U \subseteq X$, the notation $\mu(U)$ stands for $\vee_{x \in U} \mu(x)$. Given $A, B \in \mathcal{F}(X)$, the *union* of $A$ and $B$, denoted $A \cup B$, is defined by the membership function

$$(A \cup B)(x) = A(x) \vee B(x)$$

for all $x \in X$; the *intersection* of $A$ and $B$, denoted $A \cap B$, is given by the membership function

$$(A \cap B)(x) = A(x) \wedge B(x)$$

for all $x \in X$.

Recall that a *labeled transition system* consists of a set $S$ of states, a set $A$ of labels, a transition relation $\longrightarrow \subseteq S \times A \times S$, and an initial state $s_0 \in S$ (cf. [20]). Such a transition relation is equivalent to a multi-valued mapping from $S \times A$ to $S$, i.e., a mapping from $S \times A$ to $\mathcal{P}(S)$, the power set of $S$. If the label set is a singleton, the system is essentially unlabeled, and a simpler definition that omits the labels is possible. Fuzzy transition systems are a natural generalization of labeled transition systems and can be thought of as weighted graphs, possibly with an infinite number of vertices or edges.

*Definition 1:* A *fuzzy transition system* (FTS) is a four-tuple $\mathcal{S} = (S, A, \delta, s_0)$, where

(1) $S$ is a finite or infinite set of states,
(2) $A$ is a finite or infinite set of labels,
(3) $\delta$, the fuzzy transition function, is a mapping from $S \times A$ to $\mathcal{F}(S)$, or equivalently a fuzzy multi-valued mapping from $S \times A$ to $S$, and
(4) $s_0$, a member of $S$, is the initial state.

An FTS is said to be *finite* if both $S$ and $A$ are finite, and *infinite* otherwise. Labels can represent different things depending on the language of interest. Typical uses of labels include representing input expected, conditions that must be true to trigger the transition, or actions performed during the transition. Intuitively, if the FTS is in state $s \in S$ and the label $a \in A$ occurs, then it may go into any one of the states $s' \in S$, and the possibility degree of going into $s'$ is $\delta(s,a)(s')$. In other words, $\delta(s,a)(s') > 0$ means that there exists a transition from $s$ to $s'$ with label $a$ and possibility degree $\delta(s,a)(s')$. For clarity, we sometimes use the more suggestive notations like $s \xrightarrow{a} \mu$ and $s \xrightarrow{a|\gamma} s'$ to denote $\delta(s,a) = \mu$ and $\delta(s,a)(s') = \gamma$, respectively. In the obvious way, we can identify any fuzzy transition function $\delta : S \times A \longrightarrow \mathcal{F}(S)$ with a fuzzy relation $\longrightarrow_\delta : S \times A \times S \longrightarrow [0,1]$.

For example, Fig. 2 depicts a finite FTS $\mathcal{S} = (\{s_0, s_1, s_2, s_3, s_4\}, \{a, b, c\}, \delta_\mathcal{S}, s_0)$ and an infinite FTS $\mathcal{T} = (\{t_0, t_1, t_2, \ldots\}, \{a\}, \delta_\mathcal{T}, t_0)$, where an arc, say from $s_i$ to $s_j$ with label $a|x$ in $\mathcal{S}$, means that $\delta_\mathcal{S}(s_i, a)(s_j) = x$.

Clearly, a labeled transition system is a special FTS that has 0 and 1 as the possibility degrees. We remark that there are many probabilistic versions of labeled transition systems. Among others, discrete probabilistic transition systems introduced by Larsen and Skou [22] are the most relevant to FTSs above; they can be viewed as a special FTS that has probability distributions as the codomain of fuzzy transition function. Nevertheless, the semantics of probabilistic transition systems and FTSs are rather different: The weight of a transition in a probabilistic context reflects a frequency of occurrence, while the weight in a fuzzy context describes the membership grade (namely, uncertainty) of a target state. It is well known that probability theory is not capable of capturing uncertainty in all its manifestations.

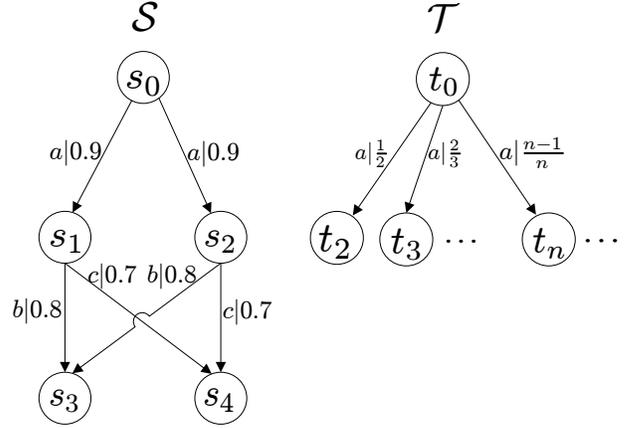

Fig. 2. A finite FTS $\mathcal{S}$ and an infinite FTS $\mathcal{T}$.

For each $s \in S$, we associate to $s$ a *fuzzy language* $\mathcal{L}_s^\mathcal{S}$, which captures the behavior of $\mathcal{S}$ starting at $s$. Formally, $\mathcal{L}_s^\mathcal{S}$ is defined as a fuzzy subset of $A^*$, the set of all finite strings over $A$ (including the empty string $\epsilon$), and given by

$$\mathcal{L}_s^\mathcal{S}(w) = \vee_{s' \in S} \delta(s, w)(s'),$$

where $\delta(s, w)(s')$ is inductively defined as follows:

$$\delta(s, \epsilon)(s') = \begin{cases} 1, & \text{if } s' = s \\ 0, & \text{otherwise} \end{cases}$$

$$\delta(s, ua)(s') = \vee_{s'' \in S}[\delta(s, u)(s'') \wedge \delta(s'', a)(s')].$$

In particular, the fuzzy language $\mathcal{L}_{s_0}^\mathcal{S}$ associated to the initial state $s_0$ of $\mathcal{S}$ is also called the *fuzzy language* generated by $\mathcal{S}$. Let $\mathcal{S}_i = (S_i, A, \delta_i, s_{0i})$ be an FTS and $s_i \in S_i$, where $i = 1, 2$. The states $s_1$ and $s_2$ are said to be *fuzzy language equivalent* if $\mathcal{L}_{s_1}^{\mathcal{S}_1} = \mathcal{L}_{s_2}^{\mathcal{S}_2}$. For instance, if $\mathcal{S}$ and $\mathcal{T}$ are the FTSs in Fig. 2, we can get by a routine computation that

$$\begin{aligned}
\mathcal{L}_{s_0}^\mathcal{S} &= \frac{1}{\epsilon} + \frac{0.9}{a} + \frac{0.8}{ab} + \frac{0.7}{ac}, \\
\mathcal{L}_{t_0}^\mathcal{T} &= \frac{1}{\epsilon} + \frac{1}{a}, \\
\mathcal{L}_{s_1}^\mathcal{S} &= \mathcal{L}_{s_2}^\mathcal{S}.
\end{aligned}$$

For labeled transition systems, one can build an overall system by building its component transition systems first and then composing them by some operators. Therefore, compositional operators can serve the need of modular specification and verification of systems. As an example, let us focus on a parallel composition operator on FTSs, in which the synchronization occurs on a set of synchronizing labels; more operators can be found in [3], [40].

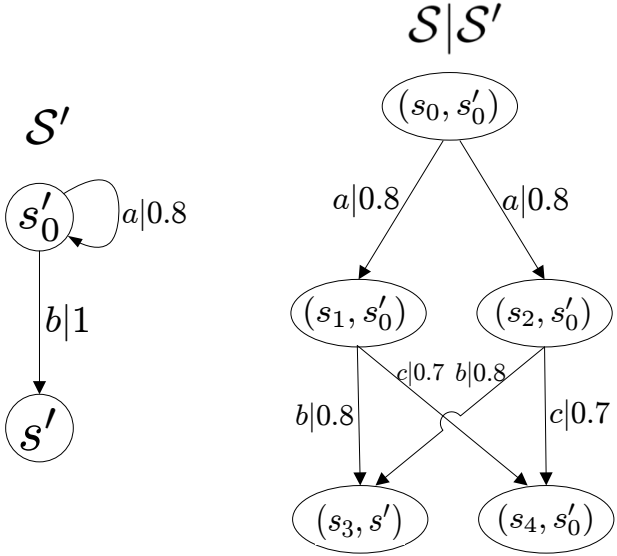

Fig. 3. FTS $\mathcal{S}'$ and the parallel composition $\mathcal{S}|\mathcal{S}'$.

Given two FTSs $\mathcal{S}_1 = (S_1, A_1, \delta_1, s_{01})$ and $\mathcal{S}_2 = (S_2, A_2, \delta_2, s_{02})$, the labels that are intended to synchronize are listed in the set $A_1 \cap A_2$ and the rest of the labels can be performed independently. More concretely, the *parallel composition* of $\mathcal{S}_1$ and $\mathcal{S}_2$ is a four-tuple

$$\mathcal{S}_1|\mathcal{S}_2 = (S_1 \times S_2, A_1 \cup A_2, \delta, (s_{01}, s_{02})),$$

where for all $(s_1, s_2), (s'_1, s'_2) \in S_1 \times S_2$ and $a \in A_1 \cup A_2$, $\delta((s_1, s_2), a)((s'_1, s'_2)) =$

$$\begin{cases} \delta_1(s_1, a)(s'_1) \wedge \delta_2(s_2, a)(s'_2), & \text{if } a \in A_1 \cap A_2 \\ \delta_1(s_1, a)(s'_1), & \text{if } a \in A_1 \backslash A_2 \text{ and } s'_2 = s_2 \\ \delta_2(s_2, a)(s'_2), & \text{if } a \in A_2 \backslash A_1 \text{ and } s'_1 = s_1 \\ 0, & \text{otherwise.} \end{cases}$$

Clearly, this constructs an FTS $\mathcal{S}_1|\mathcal{S}_2$, which represents that the systems $\mathcal{S}_1$ and $\mathcal{S}_2$ are running concurrently. The synchronization constraint $A_1 \cap A_2$ forces some labels to be carried out by both of the systems at the same time and allows all the possible interleavings of the other labels of the two systems. For example, Fig. 3 shows a parallel composition of $\mathcal{S}$ in Fig. 2 and the FTS $\mathcal{S}'$ given here.

In the literature, there are a large number of formal description tools for dynamic fuzzy systems such as various fuzzy automata [1], [8], [12], [28], [34], [41], fuzzy Petri nets [6], [9], [31], fuzzy control systems [10], [15], [30], fuzzy discrete event systems [7], [13], [25], [33], neuro-fuzzy systems [17], and so on. In general, they are not FTSs, but it is possible to translate a system's description in one of these formalisms into the FTS representing its behavior. Among others, it is perhaps the simplest for fuzzy automata since they themselves are a special class of FTSs. For comparison, we are ready to review a kind of fuzzy automata, which has been known as max-min automata in some mathematical literature [19], [35].

A fuzzy automaton is nothing other than a finite FTS with a fuzzy final state set. More precisely, it is formalized as follows.

*Definition 2:* A *fuzzy automaton* is a five-tuple $M = (Q, \Sigma, \delta, q_0, F)$, where:

(1) $Q$ is a finite set of states.
(2) $\Sigma$ is a finite input alphabet.
(3) $\delta$, the fuzzy transition function, is a function from $Q \times \Sigma$ to $\mathcal{F}(Q)$ that takes a state in $Q$ and an input symbol in $\Sigma$ as arguments and returns a fuzzy subset of $Q$.
(4) $q_0$, a member of $Q$, is the initial state.
(5) $F$ is a fuzzy subset of $Q$, called the fuzzy set of final states and for each $q \in Q$, $F(q)$ indicates intuitively the degree to which $q$ is a final state.

Denote by $\Sigma^*$ the set of all finite strings constructed by concatenation of elements of $\Sigma$, including the empty string $\epsilon$. The fuzzy language $\mathcal{L}_M$ accepted by $M$ is defined as a fuzzy subset of $\Sigma^*$ and given by

$$\mathcal{L}_M(w) = \vee_{q \in Q} [\delta(q_0, w)(q) \wedge F(q)].$$

Clearly, for any fuzzy automaton $M = (Q, \Sigma, \delta, q_0, F)$, we have that $\mathcal{L}_M \subseteq \mathcal{L}_{q_0}^{\mathcal{S}_M}$, where $\mathcal{S}_M = (Q, \Sigma, \delta, q_0)$.

## III. BISIMULATION

In the literature [3], [32], [36], the notion of bisimulation has been extended to fuzzy finite automata, in which it is an equivalence relation on the state set of the underlying automaton. In this section, we introduce a general definition of bisimulation for FTSs which is not necessarily an equivalence relation and can be applied to compare different classes of models.

To state the key definition, we need the following notion. Let $\mathcal{S}_i = (S_i, A, \delta_i, s_{0i})$, $i = 1, 2$, be an FTS. For a binary relation $R \subseteq S_1 \times S_2$, we use $\pi_1$ and $\pi_2$ for the canonical projections of $R$ on $S_1$ and $S_2$, respectively. More concretely, $\pi_1(R) = \{s \in S_1 : (s, t) \in R \text{ for some } t \in S_2\}$ and $\pi_2(R) = \{t \in S_2 : (s, t) \in R \text{ for some } s \in S_1\}$. A pair $(U, V)$ with $U \subseteq S_1$ and $V \subseteq S_2$ is called $R$-*correlational* if $\pi_1^{-1}(U) = \pi_2^{-1}(V)$, where $\pi_1^{-1}(U) = \{(s, t) \in R : s \in U\}$ and $\pi_2^{-1}(V) = \{(s, t) \in R : t \in V\}$. If $S_1 = S_2$ and $R$ is an equivalence relation, it is easy to check that $(U, V)$ is $R$-correlational if and only if $U = V = \bigcup_i C_i$ for some equivalence classes $C_i \in S_1/R$, where we are writing $S_1/R$ for the set of all equivalence classes induced by $R$.

*Definition 3:* Let $\mathcal{S}_1 = (S_1, A, \delta_1, s_{01})$ and $\mathcal{S}_2 = (S_2, A, \delta_2, s_{02})$ be two FTSs. A relation $R \subseteq S_1 \times S_2$ is called a *bisimulation* between $\mathcal{S}_1$ and $\mathcal{S}_2$ if for all $(s, t) \in R$ and $a \in A$,

(1) $s \xrightarrow{a} \mu$ in $\mathcal{S}_1$ implies $t \xrightarrow{a} \eta$ in $\mathcal{S}_2$ for some $\eta \in \mathcal{F}(S_2)$ such that $\mu(U) = \eta(V)$ for every $R$-correlational pair $(U, V)$;
(2) $t \xrightarrow{a} \eta$ in $\mathcal{S}_2$ implies $s \xrightarrow{a} \mu$ in $\mathcal{S}_1$ for some $\mu \in \mathcal{F}(S_1)$ such that $\mu(U) = \eta(V)$ for every $R$-correlational pair $(U, V)$.

Two states $s \in S_1$ and $t \in S_2$ are *bisimilar*, denoted $s \sim t$, if there exists a bisimulation $R$ between $\mathcal{S}_1$ and $\mathcal{S}_2$ such that $(s, t) \in R$. Two FTSs $\mathcal{S}_1$ and $\mathcal{S}_2$ are *bisimilar*, denoted $\mathcal{S}_1 \sim \mathcal{S}_2$, if their initial states $s_{01}$ and $s_{02}$ are bisimilar.

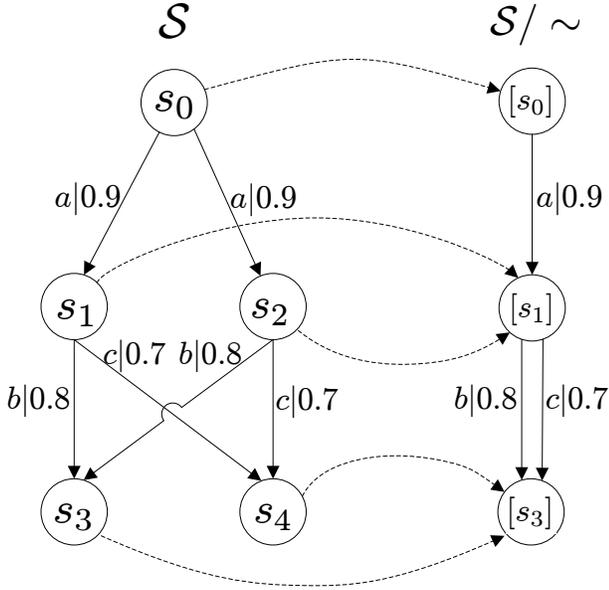

Fig. 4. Two bisimilar FTSs.

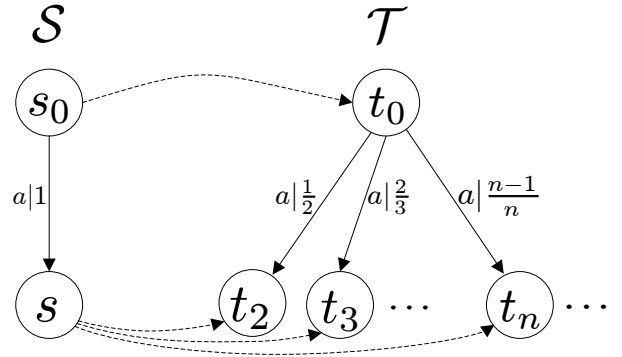

Fig. 5. Two bisimilar FTSs involving infinite states.

Intuitively, if two FTSs are bisimilar, then it is possible for each to simulate, or "track", the other's behavior. More specifically, for a relation to be a bisimulation, related states must be able to "match" transitions of each other by moving to related states with the same possibility degree. Bisimulation has a number of pleasing properties. For example, bisimilar FTSs generate the same fuzzy language, which will be proven later, and moreover, they must have the same "deadlock potential".

*Example 1:* Fig. 4 describes two bisimilar FTSs: $\mathcal{S}$ given in Fig. 2 and $\mathcal{S}/\sim$ defined here, in which we use dashed arrows to relate the states in a relation $R = \{(s_0, [s_0]), (s_1, [s_1]), (s_2, [s_1]), (s_3, [s_3]), (s_4, [s_3])\}$. By definition, it is easy to check that $R$ is a bisimulation. Therefore, $\mathcal{S}$ and $\mathcal{S}/\sim$ are bisimilar. Two other bisimilar FTSs involving infinite states are shown in Fig. 5, where $R = \{(s_0, t_0), (s, t_2), (s, t_3), \ldots\}$ gives rise to a bisimulation between the FTS $\mathcal{S}$ defined here and the FTS $\mathcal{T}$ in Fig. 2. The importance of this simple example lies in that an infinite FTS may be related to a finite one with the same behavior.

Like the notion of bisimulation in concurrency theory, bisimulation here is preserved by various operations on relations. Let $R \subseteq S \times T$ and $Q \subseteq T \times U$. Recall that the inverse $R^{-1}$ of $R$ and the composition $R \circ Q$ of $R$ and $Q$ are defined by

$$R^{-1} = \{(t, s) : (s, t) \in R\} \text{ and}$$
$$R \circ Q = \{(s, u) : \exists t \in T \text{ such that } (s, t) \in R, (t, u) \in Q\},$$

respectively.

We can now state the following proposition.

*Proposition 1:* Let $\mathcal{S}, \mathcal{S}_1, \mathcal{S}_2,$ and $\mathcal{S}_3$ be FTSs.

(1) The diagonal $\Delta_S = \{(s, s) : s \in S\}$ is a bisimulation on $\mathcal{S}$.

(2) If $R$ is a bisimulation between $\mathcal{S}_1$ and $\mathcal{S}_2$, then $R^{-1}$ is a bisimulation between $\mathcal{S}_2$ and $\mathcal{S}_1$.

(3) If $R_1$ and $R_2$ are bisimulations between $\mathcal{S}_1$ and $\mathcal{S}_2$, then so is $R_1 \cup R_2$.

(4) If $R$ is a bisimulation between $\mathcal{S}_1$ and $\mathcal{S}_2$, and $Q$ is a bisimulation between $\mathcal{S}_2$ and $\mathcal{S}_3$, then $R \circ Q$ is a bisimulation between $\mathcal{S}_1$ and $\mathcal{S}_3$.

*Proof:* See Appendix A. ∎

Let us point out that the intersection of two bisimulations is not necessarily a bisimulation. A simple counterexample is as follows: Let $\mathcal{S} = (S = \{s_0, s, t\}, A = \{a\}, \delta, s_0)$ be an FTS, where $\delta(s_0, a)(s) = \delta(s_0, a)(t) = 0.8$ and $\delta$ takes values 0 for all other cases. Consider $R_1 = \Delta_S = \{(s_0, s_0), (s, s), (t, t)\}$ and $R_2 = \{(s_0, s_0), (s, t), (t, s)\}$. It is easy to check that both $R_1$ and $R_2$ are bisimulations on $\mathcal{S}$, but their intersection $R_1 \cap R_2 = \{(s_0, s_0)\}$ fails to be a bisimulation on $\mathcal{S}$.

A bisimulation between an FTS $\mathcal{S}$ and itself is called a bisimulation *on* $\mathcal{S}$. A *bisimulation equivalence* is a bisimulation that is also an equivalence relation. As a result, a bisimulation between different FTSs must not be a bisimulation equivalence. Nevertheless, it follows from Proposition 1 that the relation $\sim$ on the set of all FTSs is an equivalence relation, that is, $\mathcal{S} \sim \mathcal{S}$, $\mathcal{S}_1 \sim \mathcal{S}_2$ implies $\mathcal{S}_2 \sim \mathcal{S}_1$, and $\mathcal{S}_1 \sim \mathcal{S}_2$ and $\mathcal{S}_2 \sim \mathcal{S}_3$ imply $\mathcal{S}_1 \sim \mathcal{S}_3$.

For bisimulation equivalence, we have the following characterization.

*Proposition 2:* Let $\mathcal{S} = (S, A, \delta, s_0)$ be an FTS. If $R$ is an equivalence relation on $S$, then the following are equivalent:

(1) $R$ is a bisimulation on $\mathcal{S}$.
(2) For any $(s, t) \in R$, $\delta(s, a)(C) = \delta(t, a)(C)$ holds for all $a \in A$ and $C \in S/R$.

*Proof:* See Appendix A. ∎

Recall that a partially ordered set $(L, \leq)$ is called a *complete lattice* if every subset of $L$ has a supremum and an infimum in $(L, \leq)$. The assertion (3) in Proposition 1 shows us a lattice structure on the set of all bisimulations between two FTSs.

*Corollary 1:* Let $\mathcal{S}_1$ and $\mathcal{S}_2$ be two FTSs. Then the set of all bisimulations between $\mathcal{S}_1$ and $\mathcal{S}_2$ is a complete lattice, with the supremum and infimum given by

$$\bigvee_i R_i = \bigcup_i R_i,$$
$$\bigwedge_i R_i = \bigcup\{R : R \text{ is a bisimulation and } R \subseteq \bigcap_i R_i\}.$$

In particular, the union of all bisimulations between $\mathcal{S}_1$ and $\mathcal{S}_2$ gives rise to the largest bisimulation, which is called a *bisimilarity* and is denoted by $\sim$. Furthermore, the largest bisimulation on one and the same FTS is an equivalence relation, as shown below.

*Corollary 2:* Let $\sim = \bigcup\{R : R \text{ is a bisimulation on } \mathcal{S}\}$. Then $\sim$ is an equivalence relation on $S$.

*Proof:* The reflexivity, symmetry, and transitivity of $\sim$ follow directly from the assertions (1), (2), and (4) in Proposition 1, respectively. ∎

The following observation shows that the operation of parallel composition is commutative.

*Proposition 3:* Let $\mathcal{S}_1 = (S_1, A_1, \delta_1, s_{01})$ and $\mathcal{S}_2 = (S_2, A_2, \delta_2, s_{02})$ be two FTSs. Then $\mathcal{S}_1|\mathcal{S}_2 \sim \mathcal{S}_2|\mathcal{S}_1$.

*Proof:* It is straightforward by verifying that $R = \{((s_1, s_2), (s_2, s_1)) : (s_1, s_2) \in S_1 \times S_2\}$ is a bisimulation between $\mathcal{S}_1|\mathcal{S}_2$ and $\mathcal{S}_2|\mathcal{S}_1$. We thus omit the details. ∎

Let us end this section with a special bisimulation for fuzzy automata (cf. [3], [32], [36]).

*Definition 4:* Let $M_i = (Q_i, \Sigma, \delta_i, q_{0i}, F_i)$, $i = 1, 2$, be a fuzzy automaton. A relation $R \subseteq Q_1 \times Q_2$ is a *bisimulation* between $M_1$ and $M_2$ if

(1) $R$ is a bisimulation between $(Q_1, \Sigma, \delta_1, q_{01})$ and $(Q_2, \Sigma, \delta_2, q_{02})$ in the sense of Definition 3;
(2) $F_1(q_1) = F_2(q_2)$ for all $(q_1, q_2) \in R$.

Two fuzzy automata $M_1$ and $M_2$ are *bisimilar* if their initial states are related by a bisimulation.

## IV. AN ALTERNATIVE DEFINITION OF BISIMILARITY

This section is devoted to an equivalent definition of bisimilarity. Based on this definition, we can explore more properties of bisimulation. In particular, as an example, we prove that bisimilarity can be preserved by parallel composition, which means that the behavior comparison of FTSs can be carried out compositionally.

Let us begin with the following observation, which provides a sufficient condition for a relation to be a bisimulation.

*Proposition 4:* Let $\mathcal{S}_1 = (S_1, A, \delta_1, s_{01})$ and $\mathcal{S}_2 = (S_2, A, \delta_2, s_{02})$ be two FTSs. A relation $R \subseteq S_1 \times S_2$ is a bisimulation between $\mathcal{S}_1$ and $\mathcal{S}_2$ if for all $(s, t) \in R$ and $a \in A$,

(1) if $s \xrightarrow{a|\gamma} s'$ for some $s' \in S_1$, then for any $\epsilon > 0$, there exists $t' \in S_2$ satisfying $t \xrightarrow{a|\gamma'} t'$, $\gamma' > \gamma - \epsilon$, and $(s', t') \in R$;

(2) if $t \xrightarrow{a|\gamma} t'$ for some $t' \in S_2$, then for any $\epsilon > 0$, there exists $s' \in S_1$ satisfying $s \xrightarrow{a|\gamma'} s'$, $\gamma' > \gamma - \epsilon$, and $(s', t') \in R$.

*Proof:* See Appendix A. ∎

For the necessity of the condition in Proposition 4, we have the following remark.

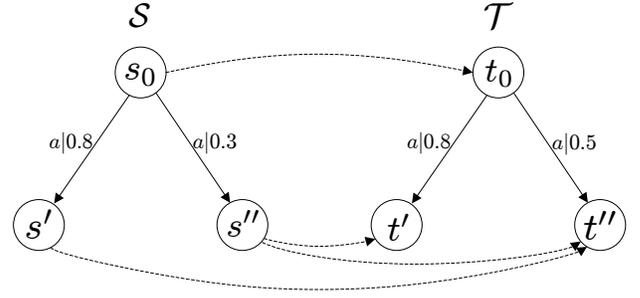

Fig. 6. A bisimulation not satisfying the conditions in Proposition 4.

*Remark 1:* In general, not all bisimulations satisfy the conditions (1) and (2) in Proposition 4. For example, consider FTSs $\mathcal{S}$ and $\mathcal{T}$ shown in Fig. 6. Taking $R = \{(s_0, t_0), (s', t''), (s'', t'), (s'', t'')\}$, it is not difficult to check that $R$ is a bisimulation by Definition 3. However, $s_0 \xrightarrow{a|0.8} s'$ in $\mathcal{S}$ cannot be matched by a transition in $\mathcal{T}$ that satisfies the condition (1) in Proposition 4.

Nevertheless, the largest bisimulation $\sim$ does satisfy the conditions (1) and (2) in Proposition 4, as shown below.

*Proposition 5:* Let $\mathcal{S}_1 = (S_1, A, \delta_1, s_{01})$ and $\mathcal{S}_2 = (S_2, A, \delta_2, s_{02})$ be two FTSs. Then for every $(s, t) \in \sim$ and $a \in A$,

(1) if $s \xrightarrow{a|\gamma} s'$ for some $s' \in S_1$, then for any $\epsilon > 0$, there exists $t' \in S_2$ satisfying $t \xrightarrow{a|\gamma'} t'$, $\gamma' > \gamma - \epsilon$, and $(s', t') \in \sim$;

(2) if $t \xrightarrow{a|\gamma} t'$ for some $t' \in S_2$, then for any $\epsilon > 0$, there exists $s' \in S_1$ satisfying $s \xrightarrow{a|\gamma'} s'$, $\gamma' > \gamma - \epsilon$, and $(s', t') \in \sim$.

*Proof:* See Appendix A. ∎

As a direct corollary of Propositions 4 and 5, we see that the following definition of bisimilarity is equivalent to the original one introduced in Definition 3.

*Definition 5:* Let $\mathcal{S}_1 = (S_1, A, \delta_1, s_{01})$ and $\mathcal{S}_2 = (S_2, A, \delta_2, s_{02})$ be two FTSs. Two states $s_1 \in S_1$ and $s_2 \in S_2$ are *bisimilar* if there exists $R \subseteq S_1 \times S_2$ such that $(s_1, s_2) \in R$, and moreover, for every $(s, t) \in R$ and $a \in A$,

(1) if $s \xrightarrow{a|\gamma} s'$ for some $s' \in S_1$, then for any $\epsilon > 0$, there exists $t' \in S_2$ satisfying $t \xrightarrow{a|\gamma'} t'$, $\gamma' > \gamma - \epsilon$, and $(s', t') \in R$;

(2) if $t \xrightarrow{a|\gamma} t'$ for some $t' \in S_2$, then for any $\epsilon > 0$, there exists $s' \in S_1$ satisfying $s \xrightarrow{a|\gamma'} s'$, $\gamma' > \gamma - \epsilon$, and $(s', t') \in R$.

In contrast with Definition 3, the above definition does not need to consider numerous $R$-correlational pairs and is more convenient to determine bisimilar states. As an example, we now prove two properties of bisimilarity by using the new definition.

Let us first show that bisimilar states are fuzzy language equivalent, as promised earlier.

*Proposition 6:* Let $\mathcal{S}_1 = (S_1, A, \delta_1, s_{01})$ and $\mathcal{S}_2 = (S_2, A, \delta_2, s_{02})$ be two FTSs. For any $s_1 \in S_1$ and $s_2 \in S_2$, if $s_1 \sim s_2$, then $\mathcal{L}_{s_1}^{\mathcal{S}_1} = \mathcal{L}_{s_2}^{\mathcal{S}_2}$.

*Proof:* See Appendix A. ∎

It follows immediately from the above proposition that bisimilar FTSs generate the same fuzzy language, and furthermore, bisimilar fuzzy automata accept the same fuzzy language. As mentioned in Introduction, we point out that the converse of Proposition 6 is not true in general, even for crisp transition systems, which is well known in concurrency theory community. For the convenience of the reader, let us record a counterexample: Consider the FTSs $\mathcal{S}$ and $\mathcal{T}$ in Fig. 1. We see by definition that

$$\mathcal{L}^{\mathcal{S}}_{s_0} = \mathcal{L}^{\mathcal{T}}_{t_0} = \frac{1}{\epsilon} + \frac{0.9}{a} + \frac{0.8}{ab} + \frac{0.7}{ac}.$$

However, it is obvious that $s_0 \not\sim t_0$ since there is no any bisimulation containing $(s_0, t_0)$. Clearly, if the FTSs under consideration are deterministic in the sense that the support of each $\delta(s,a)$ has at most one element, then the converse of Proposition 6 actually holds.

The following theorem shows that bisimilarity is preserved by the parallel composition operator defined in Section II, i.e., it is a congruence according to this operation.

*Theorem 1:* Let $\mathcal{S}_i = (S_i, A_i, \delta_{\mathcal{S}_i}, s_{0i})$ and $\mathcal{T}_i = (T_i, A_i, \delta_{\mathcal{T}_i}, t_{0i})$, where $i = 1, 2$. If $\mathcal{S}_1 \sim \mathcal{T}_1$ and $\mathcal{S}_2 \sim \mathcal{T}_2$, then $\mathcal{S}_1 | \mathcal{S}_2 \sim \mathcal{T}_1 | \mathcal{T}_2$.

*Proof:* See Appendix A. ∎

We end this section with a discussion on Definition 5. One may have noted that introducing $\epsilon$ into Definition 5 is not elegant. However, without $\epsilon$ it makes really different, as we will see below.

*Definition 6:* Let $\mathcal{S}_1 = (S_1, A, \delta_1, s_{01})$ and $\mathcal{S}_2 = (S_2, A, \delta_2, s_{02})$ be two FTSs. A relation $R \subseteq S_1 \times S_2$ is called a *strong bisimulation* between $\mathcal{S}_1$ and $\mathcal{S}_2$ if for all $(s,t) \in R$, the following conditions hold:

(1) if $s \xrightarrow{a|\gamma} s'$, then there is $t' \in S_2$ such that $t \xrightarrow{a|\gamma'} t'$ with $\gamma' \geq \gamma$ and $(s',t') \in R$;
(2) if $t \xrightarrow{a|\gamma} t'$, then there is $s' \in S_1$ such that $s \xrightarrow{a|\gamma'} s'$ with $\gamma' \geq \gamma$ and $(s',t') \in R$.

Two states $s \in S_1$ and $t \in S_2$ are said to be *strongly bisimilar*, denoted $s \simeq t$, if there exists some strong bisimulation $R$ with $(s,t) \in R$.

By definition, every strong bisimulation is a bisimulation in the sense of Definition 3 or, equivalently, Definition 5. The converse, however, does not hold. The FTSs in Fig. 5 serve as a counterexample. The relation $R = \{(s_0, t_0), (s, t_2), (s, t_3), \ldots\}$ is a bisimulation, but it is not a strong bisimulation; in particular, $s_0 \sim t_0$, but $s_0 \not\simeq t_0$. Obviously, this non-coincidence of bisimulation and strong bisimulation arises from the infinite branches of a transition. In light of this, we can provide a sufficient condition for a bisimulation to be strong. To this end, let us introduce a notion.

An FTS $\mathcal{S} = (S, A, \delta, s_0)$ is called *image finite* if for any $s \in S$ and $a \in A$, the cardinality of $\{s' \in S : \delta(s,a)(s') > 0\}$ is finite. In other words, an FTS is image finite if every state has only finitely many outgoing transitions labeled by the same label. In particular, if the state set of an FTS is finite, then the FTS must be image finite. The following result shows that the notions of bisimulation and strong bisimulation are equivalent for image finite FTSs.

*Proposition 7:* Let $\mathcal{S}_1 = (S_1, A, \delta_1, s_{01})$ and $\mathcal{S}_2 = (S_2, A, \delta_2, s_{02})$ be two image finite FTSs. Then every bisimulation $R \subseteq S_1 \times S_2$ is a strong bisimulation.

*Proof:* It follows directly from the definition of image finiteness. ∎

The above proposition, together with Definition 5, gives an easy way to verify bisimulation for image finite FTSs. In particular, it is applicable to fuzzy finite automata.

*Corollary 3:* Let $\mathcal{S}_1 = (S_1, A, \delta_1, s_{01})$ and $\mathcal{S}_2 = (S_2, A, \delta_2, s_{02})$ be two image finite FTSs. Then two states $s_1 \in S_1$ and $s_2 \in S_2$ are bisimilar if and only if there exists $R \subseteq S_1 \times S_2$ such that $(s_1, s_2) \in R$, and moreover, for all $(s,t) \in R$ and $a \in A$,

(1) if $s \xrightarrow{a|\gamma} s'$ for some $s' \in S_1$, then there exists $t' \in S_2$ satisfying $t \xrightarrow{a|\gamma'} t'$, $\gamma' \geq \gamma$, and $(s', t') \in R$;
(2) if $t \xrightarrow{a|\gamma} t'$ for some $t' \in S_2$, then there exists $s' \in S_1$ satisfying $s \xrightarrow{a|\gamma'} s'$, $\gamma' \geq \gamma$, and $(s', t') \in R$.

## V. FIXED POINT CHARACTERIZATION

In this section, we describe the bisimilarity $\sim$ as a fixed point to a suitable monotonic function. To this end, let us recall some notions and Tarski's fixed point theorem for subsequent use.

Let $(P, \preceq)$ be a partially ordered set. A function $f : P \longrightarrow P$ is said to be *monotonic* if for all $x_1, x_2 \in P$, $x_1 \preceq x_2$ implies that $f(x_1) \preceq f(x_2)$. An element $x \in P$ is called a *fixed point* of $f$ if $x = f(x)$.

The following important theorem is due to Tarski [37].

*Theorem 2 (Tarski):* Let $(P, \preceq)$ be a complete lattice and $f : P \longrightarrow P$ a monotonic function. Then $f$ has a largest fixed point $f_{\max}$ and a least fixed point $f_{\min}$ given by:

$$\begin{aligned} f_{\max} &= \bigvee \{x \in P : x \preceq f(x)\}, \\ f_{\min} &= \bigwedge \{x \in P : f(x) \preceq x\}. \end{aligned}$$

Let $\mathcal{S}_1 = (S_1, A, \delta_1, s_{01})$ and $\mathcal{S}_2 = (S_2, A, \delta_2, s_{02})$ be two FTSs. We first note that the set $\mathcal{P}(S_1 \times S_2)$ of binary relations between $\mathcal{S}_1$ and $\mathcal{S}_2$ ordered by set inclusion is a complete lattice with the set-theoretical union and intersection as the supremum and infimum, respectively.

Next we define a function $\Gamma : \mathcal{P}(S_1 \times S_2) \longrightarrow \mathcal{P}(S_1 \times S_2)$ as follows: For any $R \in \mathcal{P}(S_1 \times S_2)$, $(s,t) \in \Gamma(R)$ if and only if for any $a \in A$, the following hold:

(1) if $s \xrightarrow{a} \mu$ in $\mathcal{S}_1$, then there is $t \xrightarrow{a} \eta$ in $\mathcal{S}_2$ for some $\eta \in \mathcal{F}(S_2)$ such that $\mu(U) = \eta(V)$ for every $R$-correlational pair $(U, V)$;
(2) if $t \xrightarrow{a} \eta$ in $\mathcal{S}_2$, then there is $s \xrightarrow{a} \mu$ in $\mathcal{S}_1$ for some $\mu \in \mathcal{F}(S_1)$ such that $\mu(U) = \eta(V)$ for every $R$-correlational pair $(U, V)$.

The following proposition shows that the function $\Gamma$ defined above is monotonic.

*Proposition 8:* For any $R, R' \in \mathcal{P}(S_1 \times S_2)$, if $R \subseteq R'$, then $\Gamma(R) \subseteq \Gamma(R')$.

*Proof:* See Appendix A. ∎

As all the conditions for Tarski's theorem are satisfied, we can now characterize the bisimilarity $\sim$ as a fixed point of the function $\Gamma$.

*Theorem 3:*
(1) A relation $R \subseteq S_1 \times S_2$ is a bisimulation if and only if $R \subseteq \Gamma(R)$.
(2) The bisimilarity $\sim$ is the largest fixed point of $\Gamma$.
   *Proof:* See Appendix A. ∎

As an immediate consequence of the above theorem, we have the following.

*Corollary 4:* Define $\Gamma^0(S_1 \times S_2) = S_1 \times S_2$ and $\Gamma^{n+1}(S_1 \times S_2) = \Gamma(\Gamma^n(S_1 \times S_2))$. Then $\sim = \bigcap_{n \geq 0} \Gamma^n(S_1 \times S_2)$.

In particular, if both $S_1$ and $S_2$ are finite, then $\sim$ is equal to $\Gamma^n(S_1 \times S_2)$ for some $n \geq 0$. Note how this gives us an algorithm to calculate $\sim$ for any given finite FTSs: To compute $\sim$, simply evaluate the non-increasing sequence $\Gamma^k(S_1 \times S_2)$ for $k \geq 0$ until the sequence stabilizes. For some fuzzy finite automata, corresponding algorithms for computing $\sim$ on one and the same fuzzy finite automaton have been provided in [3], [32], [36].

## VI. SUBSYSTEMS, QUOTIENTS, AND HOMOMORPHISMS

Recall that the composition operation of FTSs seeks to build up more complex systems by combining simpler components in prescribed ways. On the contrary, the identification of subsystems and quotients is an analytic process, in which structure is to be sought in a previously existing FTS. Both subsystems and quotients are a special relationship between FTSs. More generally, it is possible to relate two FTSs via a homomorphism. This section is devoted to the notions of subsystems, quotients, and homomorphisms and their properties involving bisimulations.

Let us begin with the notion of subsystem. It describes the restrictive behavior of an FTS, in which a subset of states could be found which was closed under evolution. What that means is that states in the given subset would evolve only into each other and into no others.

*Definition 7:* Given two FTSs $S_1 = (S_1, A, \delta_1, s_{01})$ and $S_2 = (S_2, A, \delta_2, s_{02})$, we say that $S_1$ is a *subsystem* of $S_2$, written $S_1 \leq S_2$, if
(1) $S_1 \subseteq S_2$;
(2) $\delta_2(s_1, a)(s_2) = 0$ for any $s_1 \in S_1$, $a \in A$, and $s_2 \in S_2 \setminus S_1$;
(3) $\delta_1 = \delta_2|_{S_1 \times A}$.

In the above definition, the notation $\varphi|_{X'}$ means that we are restricting the mapping $\varphi$ defined on $X$ to the smaller domain $X'$.

Subsystems can be characterized in terms of bisimulations as follows.

*Proposition 9:* Let $S_i = (S_i, A, \delta_i, s_{0i})$, $i = 1, 2$, be an FTS. Then $S_1 \leq S_2$ if and only if the diagonal $\Delta_{S_1}$ of $S_1$ is a bisimulation between $S_1$ and $S_2$.
   *Proof:* See Appendix A. ∎

The notion of homomorphism is defined as follows.

*Definition 8:* Let $S_1 = (S_1, A, \delta_1, s_{01})$ and $S_2 = (S_2, A, \delta_2, s_{02})$ be two FTSs. A mapping $f : S_1 \longrightarrow S_2$ is called a *homomorphism* from $S_1$ to $S_2$ if the following hold:
(1) $f(s_{01}) = s_{02}$.
(2) $\delta_2(f(s), a)(t) = \vee\{\delta_1(s, a)(t') : t' \in S_1, f(t') = t\}$ for any $s \in S_1$, $a \in A$, and $t \in S_2$.

The *homomorphism image* of $S_1$ under a homomorphism $f$, denoted $f(S_1)$, is defined as $(f(S_1), A, \delta_2|_{f(S_1) \times A}, s_{02})$; it turns out that $f(S_1)$ is an FTS and moreover, $f(S_1) \leq S_2$. In particular, if $S_1 \leq S_2$, the embedding mapping $i : S_1 \hookrightarrow S_2$ gives rise to a homomorphism; the homomorphism image of $S_1$ under $i$ is identical to itself.

The *kernel* $\mathrm{Ker}(f)$ of a homomorphism $f : S_1 \longrightarrow S_2$ consists of all pairs of states in $S_1$ that have the same image under $f$, namely,

$$\mathrm{Ker}(f) = \{(s, s') \in S_1 \times S_1 : f(s) = f(s')\}.$$

The binary relation $\mathrm{Ker}(f)$ is a bisimulation, as shown below.

*Proposition 10:* If $f : S_1 \longrightarrow S_2$ is a homomorphism, then $\mathrm{Ker}(f)$ is a bisimulation on $S_1$.
   *Proof:* See Appendix A. ∎

Like subsystems, homomorphisms can also be characterized in terms of bisimulations.

*Theorem 4:* Let $S_1 = (S_1, A, \delta_1, s_{01})$ and $S_2 = (S_2, A, \delta_2, s_{02})$ be two FTSs and $f : S_1 \longrightarrow S_2$ a mapping. Then $f$ is a homomorphism if and only if its graph $G(f) = \{(s, f(s)) : s \in S_1\}$ is a bisimulation between $S_1$ and $S_2$ that contains $(s_{01}, s_{02})$.
   *Proof:* See Appendix A. ∎

As an immediate consequence of the above theorem, we see that $S_1 \sim S_2$ if there is a homomorphism relating them. Therefore, if the FTS $S_1$ is complex in some sense (for example, it has a large number of states), by homomorphism one may relate it to a simpler FTS without losing essential information about its behavior.

The next proposition shows that both homomorphism and its inverse preserve bisimulations.

*Proposition 11:* Let $S_1$ and $S_2$ be two FTSs and $f : S_1 \longrightarrow S_2$ a homomorphism.
(1) If $R \subseteq S_1 \times S_1$ is a bisimulation on $S_1$, then $f(R) = \{(f(s), f(s')) : (s, s') \in R\}$ is a bisimulation on $S_2$.
(2) If $R \subseteq S_2 \times S_2$ is a bisimulation on $S_2$, then $f^{-1}(R) = \{(s, s') : (f(s), f(s')) \in R\}$ is a bisimulation on $S_1$.
   *Proof:* See Appendix A. ∎

We now turn to the quotients of FTSs. According to the idea of equivalence relations, two or more states of an FTS might be regarded as being interchangeable. By aggregating interchangeable states, we get the quotient of the FTS.

*Definition 9:* Let $S = (S, A, \delta, s_0)$ be an FTS and $R$ an equivalence relation on $S$. The *quotient* of $S$ with respect to $R$ is a four-tuple $S/R = (S/R, A, \tilde{\delta}, [s_0])$, where
(1) $S/R$ is the quotient set, defined by $S/R = \{[s] : s \in S\}$ with $[s] = \{s' \in S : (s, s') \in R\}$;
(2) $\tilde{\delta} : S/R \times A \longrightarrow \mathcal{F}(S/R)$ is defined by

$$\tilde{\delta}([s], a)([s']) = \vee\{\delta(s_1, a)(s_2) : s_1 \in [s], s_2 \in [s']\}$$

for any $[s], [s'] \in S/R$ and $a \in A$.

It is easy to check that the function $\tilde{\delta}$ is well-defined. As a result, $S/R$ is an FTS. In Fig. 4, we have presented an FTS $S$ and its quotient with respect to the largest bisimulation $\sim = \{(s_0, s_0), (s_1, s_1), (s_2, s_2), (s_3, s_3), (s_4, s_4), (s_1, s_2), (s_2, s_1), (s_3, s_4), (s_4, s_3)\}$ on $S$.

*Proposition 12:* Let $\mathcal{S} = (S, A, \delta, s_0)$ be an FTS and $R$ an equivalence relation on $S$. Then $R$ is a bisimulation on $\mathcal{S}$ if and only if the quotient map $\pi : \mathcal{S} \longrightarrow \mathcal{S}/R$ gives a homomorphism from $\mathcal{S}$ to $\mathcal{S}/R$.

*Proof:* See Appendix A. ∎

Noting that usually $\mathcal{S}/R$ has significantly fewer states than $\mathcal{S}$, we may use the following result to reduce the number of states of $\mathcal{S}$ while retaining its behavior.

*Corollary 5:* Let $\mathcal{S} = (S, A, \delta, s_0)$ be an FTS and $R$ a bisimulation equivalence on $\mathcal{S}$. Then $\mathcal{S} \sim \mathcal{S}/R$.

*Proof:* If $R$ is a bisimulation equivalence on $\mathcal{S}$, then we know by Proposition 12 that the quotient map $\pi$ is a homomorphism from $\mathcal{S}$ to $\mathcal{S}/R$. It therefore follows from Theorem 4 that $\mathcal{S}$ and $\mathcal{S}/R$ are bisimilar. ∎

The best choice of $R$ in the above corollary is the bisimilarity $\sim$, the largest bisimulation equivalence on $\mathcal{S}$. Based on quotients, the following proposition gives a necessary and sufficient condition for $R = \sim$.

*Proposition 13:* Let $\mathcal{S}$ be an FTS and $R$ a bisimulation equivalence on $\mathcal{S}$. Then $R = \sim$ if and only if the diagonal $\Delta_{S/R}$ of $S/R$ is the unique bisimulation equivalence on $\mathcal{S}/R$.

*Proof:* See Appendix A. ∎

## VII. CONCLUSION

In this paper, we have introduced FTSs as a semantic model of some (possibly infinite) fuzzy systems. To compare two fuzzy systems at a finer level of "semantical sameness", bisimulation defined for fuzzy finite automata [3], [32], [36] has been extended to general FTSs. Some new characterizations of bisimulation and bisimilarity have been provided. Following the standard study on algebraic structure, we have investigated the composition operations, subsystems, quotients, and homomorphisms of FTSs. Many properties of bisimulation have been examined as well.

There are some limits and directions in which the present work can be extended. In [6], we have promised to compare the expressiveness of the two formal models of computing with words established in [6], [8] by utilizing bisimulation. In the study, we have found that it is necessary to consider bisimulation for infinite fuzzy systems since the above formal models of computing with words may have infinite labels, and moreover, we have realized that the nondeterminism in fuzzy Petri nets for computing with words [6] is different from those in fuzzy automata. This observation leads to the present paper as preliminary work along this way, and a convincing comparison of the expressiveness remains yet to be done. In addition, the paper has been concerned mainly with the theoretical development of bisimulation for FTSs. It is desirable to apply the results here to compare some practical fuzzy systems. In particular, one can use some variants of bisimilarity such as similarity to abstract from certain details of the fuzzy systems of interest. Finally, as we have seen, it is possible to relate an infinite fuzzy system to a finite one with the same behavior. It is interesting to give sufficient conditions for this kind of relevance.

## APPENDIX A

Let us first prove Proposition 4 since it makes the proof of Proposition 1 more handy.

*Proof of Proposition 4:* We only check the first condition in the definition of bisimulation, since the second one is symmetric. Suppose that $s \xrightarrow{a} \mu$ in $\mathcal{S}_1$. By the condition of the proposition, there exists an $\eta \in \mathcal{F}(S_2)$ such that $t \xrightarrow{a} \eta$. It remains to verify that $\mu(U) = \eta(V)$ for every $R$-correlational pair $(U, V)$. Without loss of generality, we assume, by contradiction, that $\mu(U) > \eta(V)$ for some $R$-correlational pair $(U, V)$. For simplicity, we write $\alpha$ and $\beta$ for $\mu(U)$ and $\eta(V)$, respectively. That is, $\alpha = \sup_{s' \in U} \mu(s')$ and $\beta = \sup_{t' \in V} \eta(t')$. Because $\alpha > \beta$ and $\alpha = \sup_{s' \in U} \mu(s')$, there is $s' \in U$ such that $s \xrightarrow{a|\gamma} s'$ with $\gamma > \alpha - \frac{\alpha - \beta}{4}$. Taking $\epsilon = \frac{\alpha - \beta}{4}$, we get by the condition (1) that there exists $t' \in S_2$ satisfying $t \xrightarrow{a|\gamma'} t'$, $\gamma' > \gamma - \epsilon$, and $(s', t') \in R$. Hence, we have that

$$\begin{aligned}
\gamma' &> \gamma - \epsilon \\
&> \alpha - \frac{\alpha - \beta}{4} - \frac{\alpha - \beta}{4} \\
&= \frac{\alpha + \beta}{2} > \beta,
\end{aligned}$$

namely, $\gamma' > \beta$. On the other hand, it follows from $(s', t') \in R$ that $\gamma' \leq \sup_{t' \in V} \eta(t') = \beta$, a contradiction. As a result, we get that $\mu(U) = \eta(V)$ for every $R$-correlational pair $(U, V)$. Consequently, $R$ is a bisimulation. ∎

To prove Proposition 1, it is convenient to have the following lemma.

*Lemma 1:* If $R \subseteq R' \subseteq S_1 \times S_2$, then every $R'$-correlational pair $(U, V)$ is also $R$-correlational.

*Proof:* Assume that $(U, V)$ is $R'$-correlational. Then we get by definition that $\pi_1'^{-1}(U) = \pi_2'^{-1}(V)$, where we are writing $\pi'$ for the canonical projections with respect to $R'$. It yields that

$$\{(s, t) \in R' : s \in U\} = \{(s, t) \in R' : t \in V\}. \quad (1)$$

For any $(s', t') \in \pi_1^{-1}(U) = \{(s, t) \in R : s \in U\}$, we see that $(s', t') \in \{(s, t) \in R' : s \in U\}$ since $R \subseteq R'$. It follows from Eq. (1) that $(s', t') \in \{(s, t) \in R' : t \in V\}$, which means that $t' \in V$. Therefore, we get that $(s', t') \in \{(s, t) \in R : t \in V\} = \pi_2^{-1}(V)$. This proves that $\pi_1^{-1}(U) \subseteq \pi_2^{-1}(V)$. By the same argument, we can prove the converse inclusion, and thus $\pi_1^{-1}(U) = \pi_2^{-1}(V)$. Hence, $(U, V)$ is $R$-correlational. ∎

*Proof of Proposition 1:* Both (1) and (2) are trivial; we only prove the assertions (3) and (4).

(3) For simplicity, let us write $R$ for $R_1 \cup R_2$. For any $(s, t) \in R$, there exists some $i \in \{1, 2\}$ such that $(s, t) \in R_i$. Therefore, if $s \xrightarrow{a} \mu$ in $\mathcal{S}_1$, then we have that $t \xrightarrow{a} \eta$ in $\mathcal{S}_2$ for some $\eta \in \mathcal{F}(S_2)$ satisfying $\mu(U) = \eta(V)$ for every $R_i$-correlational pair $(U, V)$. By Lemma 1, each $R$-correlational pair $(U, V)$ is also $R_i$-correlational. Hence, it holds that for any $R$-correlational pair $(U, V)$, $\mu(U) = \eta(V)$. Similarly, we can verify the other part in the definition of bisimulation.

(4) Let us assume that $(s, u) \in R \circ Q$. Then there is a $t \in S_2$ satisfying that $(s, t) \in R$ and $(t, u) \in Q$. Now let $s \xrightarrow{a|\gamma} s'$ and $\epsilon > 0$. Then for $\epsilon/2$, there exist $t' \in S_2$ and $\gamma'' > \gamma - \epsilon/2$ such that $t \xrightarrow{a|\gamma''} t'$ and $(s', t') \in R$. Further, for $\epsilon/2$, there exist $u' \in S_3$ and $\gamma' > \gamma'' - \epsilon/2$ such that $u \xrightarrow{a|\gamma'} u'$ and $(t', u') \in Q$. Consequently, we see that $(s', u') \in R \circ Q$ and

$$\begin{aligned} \gamma' &> \gamma'' - \epsilon/2 \\ &> \gamma - \epsilon/2 - \epsilon/2 \\ &= \gamma - \epsilon, \end{aligned}$$

as desired.

If $u \xrightarrow{a|\gamma} u'$ in $\mathcal{S}_3$, then for any $\epsilon > 0$, we can find $\gamma' > \gamma - \epsilon$ such that $s \xrightarrow{a|\gamma'} s'$ and $(s', u') \in R \circ Q$ by a similar argument. Consequently, $R \circ Q$ is a bisimulation. ∎

*Proof of Proposition 2:* Suppose that $R$ is a bisimulation on $\mathcal{S}$. Since $R$ is an equivalence relation on $S$, we see that for any $C \in S/R$, the pair $(C, C)$ is $R$-correlational. It thus follows immediately from the definition of bisimulation that (2) holds. Conversely, we only need to verify that $\delta(s, a)(U) = \delta(t, a)(V)$ for any $R$-correlational pair $(U, V)$. As $R$ is an equivalence relation, we have that $U = V = \cup_{i \in I} C_i$ for some $C_i \in S/R$. It gives rise to

$$\begin{aligned} \delta(s, a)(U) &= \delta(s, a)(\cup_{i \in I} C_i) \\ &= \vee_{i \in I} \delta(s, a)(C_i) \\ &= \vee_{i \in I} \delta(t, a)(C_i) \\ &= \delta(t, a)(\cup_{i \in I} C_i) \\ &= \delta(t, a)(V), \end{aligned}$$

namely, $\delta(s, a)(U) = \delta(t, a)(V)$, as desired. ∎

To prove Proposition 5, the notion of $z$-closure and a lemma will be handy. Following [11], we say that a relation $R \subseteq S_1 \times S_2$ is $z$-closed if for all $s, s' \in S_1$ and $t, t' \in S_2$, we have $(s, t') \in R$ whenever $(s, t) \in R$, $(s', t) \in R$, and $(s', t') \in R$. Let us set, for $n \in \mathbb{N}$, $R_0 = R$, $R_{n+1} = \{(s, t') \in S_1 \times S_2 : (s, t) \in R, (s', t) \in R_n, (s', t') \in R$ for some $s' \in S_1, t \in S_2\}$, and $R^* = \cup_{n \in \mathbb{N}} R_n$. By definition, $(s, t) \in R_n$ if and only if there are $s_0, s_1, \ldots, s_n \in S_1$ and $t_0, t_1, \ldots, t_n \in S_2$ such that $s_0 = s$, $t_n = t$, $(s_i, t_i) \in R$ for all $i \leq n$, and $(s_{i+1}, t_i) \in R$ for all $i < n$. It turns out that $R^*$ is the least $z$-closed binary relation between $S_1$ and $S_2$ that contains $R$.

*Lemma 2:* Let $\mathcal{S}_1 = (S_1, A, \delta_1)$ and $\mathcal{S}_2 = (S_2, A, \delta_2)$ be two FTSs. If $R \subseteq S_1 \times S_2$ is a bisimulation, then so is $R^*$.

*Proof:* By definition, we have that $R^* = \cup_{n \in \mathbb{N}} R_n$. Therefore, to prove the lemma, it suffices to show that for all $(s, t) \in R_n$ and $a \in A$, if $s \xrightarrow{a} \mu_s$ in $\mathcal{S}_1$, then there is a $\mu_t \in \mathcal{F}(S_2)$ such that $t \xrightarrow{a} \mu_t$ and $\mu_s(U) = \mu_t(V)$ for every $R_n$-correlational pair $(U, V)$. This can be shown by induction on $n$. For the basis step, namely, $n = 0$, it is trivial. Suppose now that $(s, t) \in R_{n+1}$. Pick $s' \in S_1$ and $t' \in S_2$ such that $(s, t') \in R$, $(s', t') \in R_n$, and $(s', t) \in R$. By the induction hypothesis and the fact that every $R$-correlational pair is necessarily $R_n$-correlational, we thus get that $t' \xrightarrow{a} \mu_{t'}$ and $s' \xrightarrow{a} \mu_{s'}$ such that $\mu_s(U) = \mu_{t'}(V) = \mu_{s'}(U) = \mu_t(V)$ for every $R_n$-correlational pair $(U, V)$. Whence, $R^*$ is a bisimulation, as desired. ∎

*Proof of Proposition 5:* Let us write $R$ for $\sim$. Since $R$ is the largest bisimulation, it follows from Lemma 2 that $R =$ $R^*$. Suppose that $(s, t) \in R$ and $s \xrightarrow{a|\gamma} s'$ for some $s' \in S_1$. Then there is $\mu \in \mathcal{F}(S_1)$ such that $\mu(s') = \gamma$. Clearly, the label $a$ can trigger the transition from $t$. Otherwise, by considering the $R$-correlational pair $(\{s'\}, \emptyset)$, we find that $R$ is not a bisimulation, which is contradictory. Therefore, there is an $\eta \in \mathcal{F}(S_2)$ such that $t \xrightarrow{a} \eta$. Set $V_0 = \{t' \in S_2 : (s', t') \in R, \eta(t') > 0\}$. Because $R$ is $z$-closed, we obtain that $V_0 \neq \emptyset$. Suppose that $(U_1, V_1)$ is the minimal $R$-correlational pair satisfying $\{s'\} \subseteq U_1$ and $V_0 \subseteq V_1$. Hence, we have that $\gamma \leq \mu(U_1) = \eta(V_1)$, and furthermore, for any $\epsilon > 0$ there exists some $t' \in V_1$ such that $t \xrightarrow{a|\gamma'} t'$ and $\gamma' > \eta(V_1) - \epsilon \geq \gamma - \epsilon$. Again, since $R$ is $z$-closed, we see that $(s', t') \in R$. Therefore, the assertion (1) in the proposition holds; the other assertion is symmetric and we thus omit its proof. ∎

*Proof of Proposition 6:* To prove that $\mathcal{L}_{s_1}^{\mathcal{S}_1} = \mathcal{L}_{s_2}^{\mathcal{S}_2}$, it is sufficient to show that $\mathcal{L}_{s_1}^{\mathcal{S}_1}(w) = \mathcal{L}_{s_2}^{\mathcal{S}_2}(w)$ for any string $w \in A^*$. We verify it by induction on the length of the string $w$.

The basis step is for strings of length 0, namely, $w = \epsilon$. In this case, $\mathcal{L}_{s_1}^{\mathcal{S}_1}(\epsilon) = 1 = \mathcal{L}_{s_2}^{\mathcal{S}_2}(\epsilon)$ by definition. Thus the basis step is true.

The induction hypothesis is that $\mathcal{L}_{s_1}^{\mathcal{S}_1}(w) = \mathcal{L}_{s_2}^{\mathcal{S}_2}(w)$ for all strings $w$ with length no more than $n$. We now prove the same for strings of the form $wa$. By contradiction, let us suppose, without loss of generality, that $\mathcal{L}_{s_1}^{\mathcal{S}_1}(wa) > \mathcal{L}_{s_2}^{\mathcal{S}_2}(wa)$ for some $a \in A$. For simplicity, we set $\alpha = \mathcal{L}_{s_1}^{\mathcal{S}_1}(wa)$ and $\beta = \mathcal{L}_{s_2}^{\mathcal{S}_2}(wa)$, that is, $\alpha = \vee_{s' \in S_1} \delta_1(s_1, wa)(s')$ and $\beta = \vee_{s'' \in S_2} \delta_2(s_2, wa)(s'')$. Note that $\frac{\alpha - \beta}{4} > 0$. By the definition of supremum, there exists an $s' \in S_1$ such that

$$\begin{aligned} \delta_1(s_1, wa)(s') &> \alpha - \frac{\alpha - \beta}{4} \\ &> \beta + \frac{\alpha - \beta}{4} \\ &\geq \delta_2(s_2, wa)(s'') + \frac{\alpha - \beta}{4}, \end{aligned}$$

namely, $\delta_1(s_1, wa)(s') > \delta_2(s_2, wa)(s'') + \frac{\alpha - \beta}{4}$ for all $s'' \in S_2$. It follows from definition that $\delta_1(s_1, wa)(s') = \vee_{s'_1 \in S_1}[\delta_1(s_1, w)(s'_1) \wedge \delta_1(s'_1, a)(s')]$ and $\delta_2(s_2, wa)(s'') = \vee_{s'_2 \in S_2}[\delta_2(s_2, w)(s'_2) \wedge \delta_1(s'_2, a)(s'')]$. Therefore, there is an $s'_1 \in S_1$ such that $\delta_1(s_1, w)(s'_1) \wedge \delta_1(s'_1, a)(s') > \delta_2(s_2, w)(s'_2) \wedge \delta_2(s'_2, a)(s'') + \frac{\alpha - \beta}{4}$ for all $s'_2, s'' \in S_2$. Using the induction hypothesis, we have that $\delta_1(s_1, w)(s'_1) \leq \vee_{s'_2 \in S_2} \delta_2(s_2, w)(s'_2)$. This forces that $\delta_1(s'_1, a)(s') > \delta_2(s'_2, a)(s'') + \frac{\alpha - \beta}{4}$ for all $s'_2, s'' \in S_2$. On the other hand, we see that $\delta_1(s_1, w)(s'_1) > 0$ from the proven fact $\delta_1(s_1, w)(s'_1) \wedge \delta_1(s'_1, a)(s') > \delta_2(s_2, w)(s'_2) \wedge \delta_2(s'_2, a)(s'') + \frac{\alpha - \beta}{4}$. It implies that there must be an $s'_2 \in S_2$ such that $s'_1 \sim s'_2$. Consider the transition $s'_1 \xrightarrow{a|\delta_1(s'_1, a)(s')} s'$ and $\epsilon = \frac{\alpha - \beta}{4}$. By the previous argument, there is no $s'' \in S_2$ such that $s'_2 \xrightarrow{a|\gamma'} s''$ and $\gamma' > \delta_1(s'_1, a)(s') - \epsilon$, which contradicts the bisimilarity of $s'_1$ and $s'_2$. The proof of the induction step is finished, as desired. ∎

*Proof of Theorem 1:* Let $R_i \subseteq S_i \times T_i$ be a bisimulation such that $\mathcal{S}_i \sim \mathcal{T}_i$. We may assume that $R_i$ satisfies the conditions of Definition 5. Consider $R = \{((s_1, s_2), (t_1, t_2)) : (s_1, t_1) \in R_1, (s_2, t_2) \in R_2\} \subseteq (S_1 \times S_2) \times (T_1 \times T_2)$. It follows that $((s_{01}, s_{02}), (t_{01}, t_{02})) \in R$. To show that $\mathcal{S}_1 | \mathcal{S}_2 \sim \mathcal{T}_1 | \mathcal{T}_2$, it suffices to verify that $R$ is a bisimulation between $\mathcal{S}_1|\mathcal{S}_2$ and $\mathcal{T}_1|\mathcal{T}_2$.

For any $((s_1, s_2), (t_1, t_2)) \in R$, if $(s_1, s_2) \xrightarrow{a|\gamma} (s'_1, s'_2)$, we now show that for any $\epsilon > 0$, there are $\gamma' > \gamma - \epsilon$ and $(t'_1, t'_2) \in T_1 \times T_2$ satisfying that $(t_1, t_2) \xrightarrow{a|\gamma'} (t'_1, t'_2)$ and $((s'_1, s'_2), (t'_1, t'_2)) \in R$. By the definition of parallel composition, three cases need to be considered.

Case 1: $a \in A_1 \cap A_2$. In this case, since $(s_1, s_2) \xrightarrow{a|\gamma} (s'_1, s'_2)$, there exists $\gamma_i \geq \gamma$ such that $s_i \xrightarrow{a|\gamma_i} s'_i$ for $i = 1, 2$. This forces that there exist $\gamma'_i > \gamma_i - \epsilon$ and $t'_i \in T_i$ such that $t_i \xrightarrow{a|\gamma'_i} t'_i$ and $(s'_i, t'_i) \in R_i$ for $i = 1, 2$. Taking $\gamma' = \gamma'_1 \wedge \gamma'_2$, we see that $\gamma' > \gamma - \epsilon$, $(t_1, t_2) \xrightarrow{a|\gamma'} (t'_1, t'_2)$, and $((s'_1, s'_2), (t'_1, t'_2)) \in R$, as desired.

Case 2: $a \in A_1 \setminus A_2$. In this case, we see that $s_1 \xrightarrow{a|\gamma} s'_1$ and $s'_2 = s_2$. Hence, there are $\gamma' > \gamma - \epsilon$ and $t'_1 \in T_1$ satisfying that $t_1 \xrightarrow{a|\gamma'} t'_1$ and $(s'_1, t'_1) \in R_1$. We thus have that $(t_1, t_2) \xrightarrow{a|\gamma'} (t'_1, t_2)$ and $((s'_1, s_2), (t'_1, t_2)) \in R$.

Case 3: $a \in A_2 \setminus A_1$. This is analogous to Case 3; we omit it.

In a similar way, we can show that if $(t_1, t_2) \xrightarrow{a|\gamma} (t'_1, t'_2)$, then for any $\epsilon > 0$, there are $\gamma' > \gamma - \epsilon$ and $(s'_1, s'_2) \in S_1 \times S_2$ satisfying that $(s_1, s_2) \xrightarrow{a|\gamma'} (s'_1, s'_2)$ and $((s'_1, s'_2), (t'_1, t'_2)) \in R$. This completes the proof of the theorem. ∎

*Proof of Proposition 8:* Suppose that $(s, t) \in \Gamma(R)$ and $s \xrightarrow{a} \mu$ in $\mathcal{S}_1$. Then by the definition of $\Gamma(R)$ there exists $t \xrightarrow{a} \eta$ in $\mathcal{S}_2$ such that $\mu(U) = \eta(V)$ for every $R$-correlational pair $(U, V)$. As $R \subseteq R'$, it follows from Lemma 1 that $\mu(U) = \eta(V)$ for every $R'$-correlational pair $(U, V)$. Consequently, $(s, t) \in \Gamma(R')$, which means that $\Gamma(R) \subseteq \Gamma(R')$, as desired. ∎

*Proof of Theorem 3:* The assertion (1) is simply a reformulation of the definition of bisimulation.

For (2), we first show that $\sim$ is a fixed point of $\Gamma$, that is, $\Gamma(\sim) = \sim$. Because $\sim$ is a bisimulation, we obtain by (1) that $\sim \subseteq \Gamma(\sim)$. It therefore follows from the monotony of $\Gamma$ that $\Gamma(\sim) \subseteq \Gamma(\Gamma(\sim))$, which means that $\Gamma(\sim)$ is a bisimulation. This forces that $\Gamma(\sim) \subseteq \sim$, since $\sim$ is the largest bisimulation. Consequently, we have that $\Gamma(\sim) = \sim$. Furthermore, $\sim$ must be the largest fixed point of $\Gamma$, since by (1) any fixed point of $\Gamma$ is a bisimulation and $\sim$ is the largest bisimulation. This completes the proof of (2). ∎

*Proof of Proposition 9:* For the necessity, suppose that $\mathcal{S}_1 \leq \mathcal{S}_2$. Hence, $S_1 \subseteq S_2$. It follows that $\Delta_{S_1} = \{(s, s) : s \in S_1\} \subseteq S_1 \times S_1 \subseteq S_1 \times S_2$. For any $(s, s) \in \Delta_{S_1}$ and $a \in A$, if $s \xrightarrow{a|\gamma} t$ in $\mathcal{S}_1$, it forces that $s \xrightarrow{a|\gamma} t$ in $\mathcal{S}_2$ since $\delta_1 = \delta_2|_{S_1 \times A}$. Clearly, $(t, t) \in \Delta_{S_1}$. Conversely, if $s \xrightarrow{a|\gamma} t$ in $\mathcal{S}_2$, we see that $t \in S_1$ because $\delta_2(s, a)(t) = 0$ for every $t \in S_2 \setminus S_1$ by definition. Therefore, $(t, t) \in \Delta_{S_1}$ in this case. Again, since $\delta_1 = \delta_2|_{S_1 \times A}$, we have that $s \xrightarrow{a|\gamma} t$ in $\mathcal{S}_1$. Whence, $\Delta_{S_1}$ is a bisimulation between $\mathcal{S}_1$ and $\mathcal{S}_2$.

We now consider the sufficiency. Assume that $\Delta_{S_1}$ is a bisimulation between $\mathcal{S}_1$ and $\mathcal{S}_2$. It follows that $S_1 \subseteq S_2$ by $\Delta_{S_1} \subseteq S_1 \times S_2$. To prove that $\delta_2(s_1, a)(s_2) = 0$ for any $s_1 \in S_1$, $a \in A$, and $s_2 \in S_2 \setminus S_1$, we suppose, by contradiction, that there exist $s \in S_1$, $a \in A$, and $t \in S_2 \setminus S_1$ such that $\delta_2(s, a)(t) > 0$. As $t \notin S_1$, the transition $s \xrightarrow{a|\delta_2(s,a)(t)} t$ in $\mathcal{S}_2$ cannot be matched by any transition in $\mathcal{S}_1$. Therefore, $\Delta_{S_1}$ is not a bisimulation, which contradicts the hypothesis. It remains to verify that $\delta_1 = \delta_2|_{S_1 \times A}$. By contradiction, assume that there are $s, t \in S_1$ and $a \in A$ such that $\delta_1(s, a)(t) > \delta_2|_{S_1 \times A}(s, a)(t) = \delta_2(s, a)(t)$ or conversely, $\delta_2|_{S_1 \times A}(s, a)(t) = \delta_2(s, a)(t) > \delta_1(s, a)(t)$. If the former inequality holds, then we can find that the transition $s \xrightarrow{a|\delta_1(s,a)(t)} t$ in $\mathcal{S}_1$ cannot be matched by any transition in $\mathcal{S}_2$, while if the latter holds, we see that the transition $s \xrightarrow{a|\delta_2(s,a)(t)} t$ in $\mathcal{S}_2$ cannot be matched by any transition in $\mathcal{S}_1$. In either case, it contradicts the assumption that $\Delta_{S_1}$ is a bisimulation. Therefore, $\mathcal{S}_1 \leq \mathcal{S}_2$. This completes the proof of the proposition. ∎

*Proof of Proposition 10:* It is easy to see that $\text{Ker}(f)$ is an equivalence relation on $S_1$. For any $(s, s') \in \text{Ker}(f)$, $a \in A$, and $C \in S_1/\text{Ker}(f)$, we obtain by the definition of homomorphism that

$$\begin{aligned}
\delta_1(s, a)(C) &= \vee_{t \in C} \delta_1(s, a)(t) \\
&= \delta_2(f(s), a)(f(t)) \\
&= \delta_2(f(s'), a)(f(t)) \\
&= \vee_{t \in C} \delta_1(s', a)(t) \\
&= \delta_1(s', a)(C).
\end{aligned}$$

It therefore follows from Proposition 2 that $\text{Ker}(f)$ is a bisimulation on $\mathcal{S}_1$, finishing the proof. ∎

To give the proof of Theorem 4, it is convenient to have the following lemma.

*Lemma 3:* Let $\mathcal{S}_1 = (S_1, A, \delta_1, s_{01})$ and $\mathcal{S}_2 = (S_2, A, \delta_2, s_{02})$ be two FTSs. Suppose that $R$ is a bisimulation between $\mathcal{S}_1$ and $\mathcal{S}_2$. Then for any $(s, t) \in R$ and $a \in A$, we have the following:

(1) $\delta_1(s, a)(s') \leq \vee\{\delta_2(t, a)(t') : t' \in S_2, (s', t') \in R\}$ for any $s' \in S_1$.
(2) $\delta_2(t, a)(t') \leq \vee\{\delta_1(s, a)(s') : s' \in S_1, (s', t') \in R\}$ for any $t' \in S_2$.

*Proof:* (1) and (2) are symmetric, so we only prove (1). Set $\gamma = \delta_1(s, a)(s')$ and $\gamma_0 = \vee\{\delta_2(t, a)(t') : t' \in S_2, (s', t') \in R\}$. By contradiction, assume that $\gamma > \gamma_0$. Then taking $\epsilon = \gamma - \gamma_0$, there exist $\gamma' > \gamma - \epsilon = \gamma_0$ and $t' \in S_2$ such that $t \xrightarrow{a|\gamma'} t'$ and $(s', t') \in R$. We thus see that $\gamma' \leq \gamma_0$, a contradiction. Hence, $\gamma \leq \gamma_0$, as desired. This completes the proof of (1). ∎

*Proof of Theorem 4:* We prove the necessity first. Suppose that $f : S_1 \to S_2$ is a homomorphism. Then we have that $(s_{01}, s_{02}) \in G(f)$. For any $(s, f(s)) \in G(f)$ and $s \xrightarrow{a|\gamma} s'$, we get by the definition of homomorphism that $\delta_2(f(s), a)(f(s')) = \vee\{\delta_1(s, a)(s'') : s'' \in S_1, f(s'') = f(s')\}$. Taking $\gamma' = \delta_2(f(s), a)(f(s'))$, we see that $\gamma' > \gamma - \epsilon$ for any $\epsilon > 0$, $f(s) \xrightarrow{a|\gamma'} f(s')$, and $(s', f(s')) \in G(f)$. Conversely, if $f(s) \xrightarrow{a|\gamma} t$, then $\gamma = \delta_2(f(s), a)(t) = \vee\{\delta_1(s, a)(s'') : s'' \in S_1, f(s'') = t\}$ by the definition of homomorphism. Thus, for any $\epsilon > 0$, there exists $s' \in S_1$ such that $f(s') = t$ and $\delta_1(s, a)(s') > \gamma - \epsilon$. That is, for any $\epsilon > 0$, there exist $\gamma' = \delta_1(s,a)(s') > \gamma - \epsilon$ and $s' \in S$ such that $s \xrightarrow{a|\gamma'} s'$ and $(s',t) \in G(f)$. So $G(f)$ is a bisimulation between $\mathcal{S}_1$ and $\mathcal{S}_2$.

Next, to see the sufficiency, assume that $G(f)$ is a bisimulation containing $(s_{01}, s_{02})$. By definition, we need to check that $\delta_2(f(s),a)(t) = \vee\{\delta_2(s,a)(s') : s' \in S_1, f(s') = t\}$ for any $s \in S_1$, $a \in A$, and $t \in S_2$. Let $\gamma_0 = \vee\{\delta_1(s,a)(s') : s' \in S_1, f(s') = t\}$ and $\gamma_1 = \delta_2(f(s),a)(t)$. Then we have that $\gamma_1 \leq \gamma_0$ by Lemma 3. On the other hand, for any $s' \in S_1$ with $f(s') = t$, we get, again by Lemma 3, that

$$\begin{aligned}\delta_1(s,a)(s') &\leq \vee\{\delta_2(f(s),a)(t') : t' = f(s')\} \\ &= \delta_2(f(s),a)(t) \\ &= \gamma_1,\end{aligned}$$

which means that $\gamma_0 \leq \gamma_1$. Hence, $\gamma_1 = \gamma_0$, finishing the proof of the theorem. ∎

*Proof of Proposition 11:* (1) Let $(f(s), f(s')) \in f(R)$ and $f(s) \xrightarrow{a|\gamma} t$. We now show that for any $\epsilon > 0$, there exist $\gamma' > \gamma - \epsilon$ and $t' \in S$ satisfying that $f(s') \xrightarrow{a|\gamma'} f(t')$ and $(t, f(t')) \in f(R)$. Since $f$ is a homomorphism, there exists $s'' \in S_1$ such that $t = f(s'')$. By Theorem 4, we know that $G(f) = \{(s, f(s)) : s \in S_1\}$ is a bisimulation. Thus, there are $\gamma_1 > \gamma - \epsilon/3$ and $s_1 \in S_1$ such that $s \xrightarrow{a|\gamma_1} s_1$ and $f(s_1) = t$. It follows from $(f(s), f(s')) \in f(R)$ that $(s, s') \in R$. Therefore, there exist $\gamma_2 > \gamma_1 - \epsilon/3$ and $t' \in S_1$ such that $s' \xrightarrow{a|\gamma_2} t'$ and $(s_1, t') \in R$. Again, using the fact that $G(f)$ is a bisimulation, there exists $\gamma' > \gamma_2 - \epsilon/3$ such that $f(s') \xrightarrow{a|\gamma'} f(t')$. Moreover, we have that $\gamma' > \gamma - \epsilon$ and $(t, f(t')) = (f(s_1), f(t')) \in f(R)$. Similarly, we can prove that if $f(s') \xrightarrow{a|\gamma} t$, then for any $\epsilon > 0$, there exist $\gamma' > \gamma - \epsilon$ and $t' \in S_1$ satisfying that $f(s) \xrightarrow{a|\gamma'} f(t')$ and $(t, f(t')) \in f(R)$. The proof of (1) is completed.

(2) It can be proved by imitating the proof of (1). ∎

*Proof of Proposition 12:* Suppose that $R$ is a bisimulation on $\mathcal{S}$. By definition, we see that $\pi(s_0) = [s_0]$. For any $s, s' \in S$ and $a \in A$, it follows from Proposition 2 that

$$\delta(s,a)([s']) = \delta(s'',a)([s'])$$

for all $s'' \in [s]$, which means that

$$\delta(s,a)([s']) = \vee_{s'' \in [s]} \delta(s'',a)([s']).$$

This forces that

$$\tilde{\delta}([s],a)([s']) = \delta(s,a)([s']) = \vee\{\delta(s,a)(t') : t' \in [s']\}.$$

Therefore, $\pi$ is a homomorphism from $\mathcal{S}$ to $\mathcal{S}/R$ by definition.

Conversely, if the quotient map $\pi : S \longrightarrow S/R$ gives a homomorphism from $\mathcal{S}$ to $\mathcal{S}/R$, then by Proposition 10 the kernel $\text{Ker}(\pi) = \{(s,s') \in S \times S : [s] = [s']\}$ of $\pi$ is a bisimulation on $\mathcal{S}$. Observe that $\text{Ker}(\pi) = R$, so $R$ is a bisimulation on $\mathcal{S}$, which completes the proof of the proposition. ∎

*Proof of Proposition 13:* We first show the necessity. By contradiction, suppose that $Q$ is another bisimulation equivalence on $\mathcal{S}/R$. Then there is $([s],[s']) \in Q \setminus \Delta_{S/R}$, which means that $[s] \neq [s']$, i.e., $(s,s') \notin R$. Since $R$ is a bisimulation equivalence on $\mathcal{S}$, we get by Proposition 12 that the quotient map $\pi : S \longrightarrow S/R$ is a homomorphism from $\mathcal{S}$ to $\mathcal{S}/R$. By (2) of Proposition 11, we see that $\pi^{-1}(Q)$ is a bisimulation on $\mathcal{S}$. Since $\sim$ is the largest one, we have that $\pi^{-1}(Q) \subseteq \sim = R$. Noting that $(s,s') \in \pi^{-1}(Q)$, we find that $(s,s') \in R$, a contradiction. Hence, $\Delta_{S/R}$ is the only bisimulation equivalence on $\mathcal{S}/R$.

Now, we consider the sufficiency. Assume that $\Delta_{S/R}$ is the unique bisimulation equivalence on $\mathcal{S}/R$. Then $\Delta_{S/R}$ is also the largest bisimulation on $\mathcal{S}/R$. Let $R'$ be an arbitrary bisimulation on $\mathcal{S}$. Using Proposition 11, we know that $\pi(R')$ is a bisimulation on $\mathcal{S}/R$. So $\pi(R') \subseteq \Delta_{S/R}$, which implies that for any $(s,s') \in R'$, $[s] = [s']$. Therefore, $(s,s') \in R$, and thus $R' \subseteq R$. It forces that $R = \sim$, finishing the proof. ∎

## REFERENCES

[1] G. Bailador and G. Triviño, "Pattern recognition using temporal fuzzy automata," *Fuzzy Sets Syst.*, vol. 161, pp. 37-55, 2010.

[2] P. Buchholz, "Exact performance equivalence: an equivalence relation for stochastic automata," *Theor. Comput. Sci.*, vol. 215, pp. 263-287, 1999.

[3] P. Buchholz, "Bisimulation relations for weighted automata," *Theor. Comput. Sci.*, vol. 393, pp. 109-123, 2008.

[4] A. J. Bugarin and S. Barro, "Fuzzy reasoning supported by Petri nets," *IEEE Trans. Fuzzy Syst.*, vol. 2, pp. 135-150, 1994.

[5] T. Cao and A. C. Sanderson, "Task sequence planning using fuzzy Petri nets," *IEEE Trans. Syst., Man, Cybern.*, vol. 25, pp. 755-768, May 1995.

[6] Y. Cao and G. Chen, "A fuzzy Petri-nets model for computing with words," *IEEE Trans. Fuzzy Syst.*, vol. 18, pp. 486-499, 2010.

[7] Y. Cao and M. S. Ying, "Supervisory control of fuzzy discrete event systems," *IEEE Trans. Syst., Man, Cybern., Part B*, vol. 35, pp. 366-371, Apr. 2005.

[8] Y. Cao, M. Ying, and G. Chen, "Retraction and generalized extension of computing with words," *IEEE Trans. Fuzzy Syst.*, vol. 15, pp. 1238-1250, Dec. 2007.

[9] J. Cardoso, R. Valette, and D. Dubois, "Fuzzy Petri net: an overview," in *Proc. 13th IFAC World Congr.*, San Francisco, CA, June 30-July 5, 1996, pp. 443-448.

[10] Y. Y. Chen and T. C. Tsao, "A description of the dynamic behavior of fuzzy systems," *IEEE Trans. Syst., Man, Cybern.*, vol. 19, pp. 745-755, 1989.

[11] E. P. de Vink and J. J. M. M. Rutten, "Bisimulation for probabilistic transition systems: a coalgebraic approach," *Theor. Comput. Sci.*, vol. 221, pp. 271-293, 1999.

[12] M. Doostfatemeh and S. C. Kremer, "New directions in fuzzy automata," *Int. J. Approx. Reason.*, vol. 38, pp. 175-214, 2005.

[13] X. Du, H. Ying, and F. Lin, "Theory of extended fuzzy discrete-event systems for handling ranges of knowledge uncertainties and subjectivity," *IEEE Trans. Fuzzy Syst.*, vol. 17, pp. 316-328, 2009.

[14] D. Dubois and H. Prade, *Fuzzy Sets and Systems: Theory and Applications.* New York: Academic, 1980.

[15] G. Feng, "A survey on analysis and design of model-based fuzzy control systems," *IEEE Trans. Fuzzy Syst.*, vol. 14, pp. 676-697, 2006.

[16] R. J. van Glabbeek, S. A. Smolka, and B. Steffen, "Reactive, generative, and stratified models of probabilistic processes" *Inf. Comput.*, vol. 121, no. 1, pp. 59-80, 1995.

[17] C. L. Giles, C. W. Omlin, and K. K. Thornber, "Equivalence in knowledge representation: automata, recurrent neural networks, and dynamical fuzzy systems," *Proc. IEEE*, vol. 87, pp. 1623-1640, 1999.

[18] X. G. He, "Fuzzy Petri net," *Chinese J. Comput.*, vol. 17, no. 12, pp. 946-950, 1994, (in Chinese).

[19] A. Kandel and S. C. Lee, *Fuzzy Switching and Automata: Theory and Applications.* New York: Russak, 1979.

[20] R. M. Keller, "Formal verification of parallel programs," *Comm. ACM*, vol. 19, pp. 371-384, 1976.

[21] M. Kurano, M. Yasuda, J. Nakagami, and Y. Yoshida, "A limit theorem in some dynamic fuzzy systems", *Fuzzy Sets Syst.*, vol. 51, pp. 83-88, 1992.

[22] K. G. Larsen and A. Skou, "Bisimulation through probabilistic testing," *Inf. Comput.*, vol. 94, pp. 1-28, 1991.

[23] Y. M. Li, "Fuzzy Turing machines: variants and universality," *IEEE Trans. Fuzzy Syst.*, vol. 16, no. 6, pp. 1491-1502, 2008.

[24] Y. M. Li, "Approximation and robustness of fuzzy finite automata," *Int. J. Approx. Reason.*, vol. 47, no. 2, pp. 247-257, 2008.

[25] F. Lin and H. Ying, "Modeling and control of fuzzy discrete event systems," *IEEE Trans. Syst., Man, Cybern., Part B*, vol. 32, pp. 408-415, Aug. 2002.

[26] N. López and M. Núñez, "An overview of probabilistic process algebras and their equivalences," in: *C. Baier et al. (Eds.), Validation of Stochastic Systems, Lect. Notes Comput. Sci., vol. 2925,* Springer, Berlin, 2004, pp. 89-123.

[27] R. Milner, *Communication and Concurrency*. Upper Saddle River, NJ: Boca Raton, FL: Prentice-Hall, 1989.

[28] J. N. Mordeson and D. S. Malik, *Fuzzy Automata and Languages: Theory and Applications*. Boca Raton, FL: Chapman & Hall/CRC, 2002.

[29] D. Park, "Concurrency and automata on infinite sequences," in: *G. Goos and J. Hartmanis (Eds.), Proc. 5th GI Conference on Theoretical Computer Science, Lect. Notes Comput. Sci., vol. 104,* Springer, Berlin, 1981, pp. 167-183.

[30] W. Pedrycz, *Fuzzy Control and Fuzzy Systems*. John Wiley & Sons, Inc., New York, NY, USA, 1993.

[31] W. Pedrycz and F. Gomide, "A generalized fuzzy Petri net model," *IEEE Trans. Fuzzy Syst.*, vol. 2, no. 4, pp. 295-301, Nov. 1994.

[32] T. Petković, "Congruences and homomorphisms of fuzzy automata," *Fuzzy Sets Syst.*, vol. 157, pp. 444-458, 2006.

[33] D. W. Qiu, "Supervisory control of fuzzy discrete event systems: a formal approach," *IEEE Trans. Syst., Man, Cybern., Part B*, vol. 35, pp. 72-88, Feb. 2005.

[34] G. G. Rigatos, "Fuzzy stochastic automata for intelligent vehicle control," *IEEE Trans. Industr. Electr.*, vol. 50, pp. 76-79, 2003.

[35] E. S. Santos, "Maxmin automata," *Inform. Contr.*, vol. 13, pp. 363-377, 1968.

[36] D. D. Sun, Y. M. Li, and W. W. Yang, "Bisimulation relations for fuzzy finite automata," *Fuzzy Syst. Math.*, vol. 23, pp. 92-100, 2009 (in Chinese).

[37] A. Tarski, "A lattice-theoretical fixpoint theorem and its applications," *Pac. J. Math.*, vol. 5, pp. 285-309, 1955.

[38] W. G. Wee, *On generalizations of adaptive algorithm and application of the fuzzy sets concept to pattern classification*, Ph.D. Thesis, Purdue University, 1967.

[39] W. G. Wee and K. S. Fu, "A formulation of fuzzy automata and its application as a model of learning systems," *IEEE Trans. Syst. Sci. Cybern.*, vol. 5, no. 3, pp. 215-223, 1969.

[40] G. Winskel and M. Nielsen, "Models for concurrency," in: *S. Abramsky, Dov M. Gabby, T. S. E. Maibaum (Eds.), Handbook of Logic in Computer Science, vol. 4: Semantic Modelling,* Clarendon Press, Oxford, 1995, pp. 1-139.

[41] J. Virant and N. Zimic, "Fuzzy automata with fuzzy relief," *IEEE Trans. Fuzzy Syst.*, vol. 3, no. 1, pp. 69-74, Feb. 1995.

[42] L. A. Zadeh, "Fuzzy sets and systems", in *Proc. Symp. Syst. Theory*, Polytech. Inst. Brooklyn, New York, 1965, pp. 29-37.